\documentclass{article}

\usepackage{microtype}
\usepackage{graphicx}
\usepackage{subcaption}
\usepackage{booktabs} % for professional tables
\usepackage{multirow}
\usepackage{tabularx}
\newcolumntype{C}{>{\centering\arraybackslash}X}

\usepackage{xurl}
\usepackage{hyperref}

\usepackage[accepted]{icml2026}

\usepackage{amsmath}
\usepackage{amssymb}
\usepackage{mathtools}
\usepackage{amsthm}
\usepackage{enumitem}

\usepackage[capitalize,noabbrev]{cleveref}

\theoremstyle{plain}

\theoremstyle{definition}

\theoremstyle{remark}

\usepackage[textsize=tiny]{todonotes}

\icmltitlerunning{InverseScope: Scalable Activation Inversion for Interpreting LLMs}

\begin{document}

\twocolumn[
  \icmltitle{InverseScope: Scalable Activation Inversion for Interpreting LLMs}
  % From Atoms to Trees: Building a Structured Feature Forest with Hierarchical SAEs




  \begin{icmlauthorlist}
    \icmlauthor{Yifan Luo}{sms}
    \icmlauthor{Zhennan Zhou}{sswu}
    \icmlauthor{Bin Dong}{bicmr,cmlr}
  \end{icmlauthorlist}

  \icmlaffiliation{sms}{School of Mathematical Science, Peking University, Beijing, China}
  \icmlaffiliation{bicmr}{Beijing International Center for Mathematical Research and the New Cornerstone
Science Laboratory, Peking University, Beijing, China}
  \icmlaffiliation{cmlr}{Center for Machine Learning Research, Peking University, Beijing, China}
  \icmlaffiliation{sswu}{School of Science, Westlake University, Hangzhou, China}

  \icmlcorrespondingauthor{Yifan Luo}{luoyf@pku.edu.cn}
  \icmlcorrespondingauthor{Bin Dong}{dongbin@math.pku.edu.cn}

  \icmlkeywords{ICML, Mechanism Interpretability}

  \vskip 0.3in
]



\printAffiliationsAndNotice{}  % no special notice (required even if empty)

\begin{abstract}
Understanding the internal representations of large language models (LLMs) is a central challenge in interpretability research. Existing feature interpretability methods often rely on strong structural assumptions—such as linearity or sparsity—that may not hold in practice. In this work, we introduce InverseScope, an assumption-light and scalable framework for interpreting neural activations via input inversion. Given a target activation, InverseScope characterizes its encoded information by generating natural-language inputs that produce nearby activations, grounding abstract internal states in concrete language. To overcome the prohibitive cost of sampling in high-dimensional activation spaces, we propose a novel control-layer conditioning architecture that substantially improves sample efficiency compared to prior token-prepending approaches. We demonstrate that InverseScope reveals rich geometric structure in LLM representation spaces, including sentence-level linear analogies. The framework scales to state-of-the-art open-source models of up to 14B parameters and generalizes to out-of-distribution inputs, enabling systematic analysis of activation neighborhoods.
\end{abstract}

\section{Introduction}

Recent advances in mechanistic interpretability aim to reverse-engineer neural networks' computations into human-understandable processes \citep{bereska2024mechanistic, sharkey2025open}. A central task in this field is feature interpretability, which seeks to understand what information is encoded in a network's activations and how it is represented. This understanding is crucial for analyzing how information propagates and is processed across layers. Numerous methods have been proposed for feature interpretability, including linear probing \citep{Alain2016UnderstandingIL, Park2023TheLR}, sparse dictionary learning \citep{cunningham2023sparse, gao2024scaling}, and other approaches \citep{bau2017network, xu2024uncovering}.

Despite their successes, these methods share a fundamental limitation: they rely on strong assumptions about the structure of neural representations. Specifically, linear probing assumes a linear relationship between activations and specific concepts in the inputs, and sparse dictionary learning presupposes that activations can be decomposed into a sparse sum of linear directions. The validity of these assumptions remains an open and actively debated question, particularly in the context of LLMs \citep{levy2024language, engelsnot, Smith2024StrongFeatureHypothesis}. Designing experiments that rigorously test these hypotheses is itself a challenging problem, making it difficult to assess the reliability of interpretations derived from such approaches.

% These limitations highlight the need for interpretability methods that rely on minimal assumptions about the structure of neural representations. One promising direction is to invert activation back to the input space, where human intuitions are more naturally grounded. The strategy of connecting activations back to the inputs that produce them has a long-standing history in interpretability research, including early work on activation maximization \citep{erhan2009visualizing, Nguyen2016MultifacetedFV} and neural representation inversion \citep{Mahendran2014UnderstandingDI}. Building on this line, InversionView \citep{huang2024inversionview} adapts the idea of previous inversion-based methods to language models by interpreting the information encoded in an activation through the collection of inputs that generates similar representations. These methods enable feature interpretability by relying only on the geometric proximity of activations, without presupposing restrictive structural assumptions like linearity or sparsity.

To address these limitations, a promising alternative is to interpret activations by inverting them back to the input space, where representations are naturally grounded in human-understandable language. This inversion-based paradigm has a long history in computer vision \citep{erhan2009visualizing, Mahendran2014UnderstandingDI} and has recently gained traction in the NLP community. For instance, Rep2Text \citep{dong2025rep2text} demonstrates that the representation at a single last-token position encodes sufficient information to autoregressively reconstruct the full input text via a trained adapter, while InversionView \citep{huang2024inversionview} shows that the information encoded in a hidden state can be characterized by collecting a distribution of inputs that yield similar activations. Together, these works highlight the potential of inversion-based methods to provide an assumption-light and intuitive framework for feature interpretability, enabling a direct mapping between abstract internal states and concrete natural language.

However, existing inversion methods each face their own limitations. Rep2Text \citep{dong2025rep2text} targets exact reconstruction of the original input, and consequently lacks the neighborhood perspective central to interpretability. InversionView \citep{huang2024inversionview} addresses this gap by framing inversion as neighborhood characterization, but its applicability has been confined to task-specific templates and small-scale models, leaving the general-distribution setting unexplored. Critically, all three methods share the same architectural family: the target activation is projected into the token embedding space and placed as a soft token alongside the other input embeddings. Our ablation experiments demonstrate that this input-level conditioning is architecturally suboptimal: conditioning at intermediate layers via dedicated control layers yields substantially better sample efficiency (Section~\ref{sec:retrieval_efficiency}).

Our main contributions are as follows:
\begin{itemize}
\item \textbf{Control-layer conditioning architecture.} We introduce a conditional generation architecture in which control layers inject the target activation $z$ into the backbone at every Transformer layer, analogously to cross-attention. This design substantially outperforms the token-prepending approach used in prior work in both sample efficiency and training stability, enabling reliable inversion at scale.

\item \textbf{Scaling to general distributions and large models.} We scale inversion-based interpretability to general-domain text (Wikipedia and Project Gutenberg) and to LLMs of up to 14B parameters, removing the dependence on task-specific templates that has restricted prior methods. We further characterize how inversion quality varies with target model size and generator capacity.

\item \textbf{Empirical analysis of representation geometry and safety interventions.} We demonstrate that InverseScope recovers sentence-level linear analogy structure in LLM activations and provides interpretable neighborhoods for activations produced by instruction-following prompts. We apply the framework to analyze refusal-direction ablation, showing that it renders the representational effect of the intervention directly legible in natural language.
\end{itemize}

Collectively, these contributions advance inversion-based interpretability from small-scale, task-specific settings to general-purpose analysis of state-of-the-art LLMs, and demonstrate the utility of the framework for both geometric probing and safety-relevant applications.

\section{Related works}

\paragraph{Activation interpretation methods.} 
A variety of methods have been developed to interpret neural network features. Classical approaches such as linear probing train simple classifiers on activations to identify linearly encoded features \citep{Alain2016UnderstandingIL, Park2023TheLR}. More recent work includes sparse dictionary learning, which decompose activations into sparse and interpretable components to disentangle feature representations \citep{cunningham2023sparse, gao2024scaling}. While effective, these approaches typically rely on strong structural assumptions about the underlying representations. For example, that features are linearly separable, sparse, or decomposable into a fixed dictionary. Such assumptions may not hold in general, potentially limiting their ability to capture more complex or distributed representations in large language models.

\paragraph{Natural language interpretability.} 
Several recent works have explored assigning human-interpretable labels—such as natural language descriptions—to the internal activations of LLMs. Early approaches such as the logit lens interpret intermediate activations by projecting them directly into the vocabulary space and reading out the most likely tokens, effectively providing single-word descriptions of representations \citep{geva2022transformer}. Building on this idea, training-free methods like SelfIE \citep{chen2024selfie} and PatchScope \citep{Ghandeharioun2024PatchscopesAU} use a pretrained LLM to read out the information encoded in residual stream activations. Complementing these approaches, LatentQA \citep{pan2024latentqa} trains a decoder model to answer natural language questions about activations. More recent work further generalizes this paradigm by treating activations as a new input modality: Activation Oracles are trained to answer arbitrary queries about activations, enabling more flexible and general-purpose natural language interpretation \citep{Karvonen2025ActivationOT}. Despite their flexibility, these methods face a fundamental challenge: the natural language explanations they produce are difficult to verify. In particular, they may suffer from hallucination, where the generated descriptions are fluent and plausible but do not faithfully reflect the underlying representations, making it challenging to assess their reliability.

\paragraph{Inversion-based interpretability.} 
A complementary line of research focuses on interpreting model activations by identifying the inputs that give rise to them, a strategy with a long history in computer vision via activation maximization and representation inversion \citep{erhan2009visualizing, Nguyen2016MultifacetedFV, Mahendran2014UnderstandingDI}. Recent work has extended this paradigm to language models along two distinct directions. Rep2Text \citep{dong2025rep2text} trains a learned adapter to autoregressively decode the full input text from the single last-token representation of an LLM, demonstrating that substantial lexical and semantic information is recoverable from this compressed summary vector. InversionView \citep{huang2024inversionview} takes a different approach: rather than reconstructing the exact original input, it samples a distribution of inputs whose activations are close to the target, thereby characterizing what information is encoded in a given hidden state. Concurrently, a complementary line of work proves that decoder-only Transformer LLMs are almost surely injective: different prompts produce different hidden states at any layer \citep{nikolaou2025invertible}. They operationalize this result via SipIt, a token-by-token search algorithm that provably reconstructs the exact original prompt from the full hidden-state matrix (all token positions) at any layer. InverseScope differs from all three in both interpretive goal and architecture. SipIt and Rep2Text target exact reconstruction of the original input; InversionView characterizes the activation neighborhood but is confined to task-specific settings. Architecturally, InversionView, Activation Oracles, and Rep2Text all belong to the same family: the target activation is linearly projected into the token embedding space and placed as a soft token alongside the model's input embeddings. InverseScope instead injects the conditioning signal at every Transformer layer via dedicated control layers, generalizes to arbitrary inference positions and general-domain data, and scales to LLMs of up to 14B parameters.

\paragraph{Conditional and controllable generation.} 
Conditional generation, which generates outputs based on a constrained intermediate state, is a foundational technique in deep learning. Initially established with Conditional VAEs \citep{Sohn2015LearningSO} and Conditional GANs \citep{Mirza2014ConditionalGA}, this paradigm has evolved into methods where the conditioning signal specifies desired structure in the output. Notable examples include modern text-to-image synthesis systems \citep{Ramesh2022HierarchicalTI} and structured guidance methods such as ControlNet \citep{Zhang2023AddingCC}, which condition large generative models on auxiliary inputs. Similarly, controllable text generation \citep{Keskar2019CTRLAC} steers language model outputs using explicit attributes. These developments highlight the generality of treating intermediate representations as controllable interfaces for generation.

\section{InverseScope}

In this section, we introduce our method: its intuition, network architecture, training dataset, and training procedure.

\subsection{Intuition}

Inversion-based interpretability builds on a core intuition: similar activations tend to encode semantically similar information \citep{bengio2013representation}. In this paper, we use the term “activation” to specifically denote the output of an LLM layer at a specific inference position, $z\in\mathbb{R}^d$. If two distinct inputs produce nearly identical activations, then the differences between those inputs are unlikely to be encoded in that representation. Conversely, if nearby activations consistently correspond to inputs sharing a particular pattern, this indicates that the pattern is encoded within that region of latent space. This follows from the continuity of neural networks: since downstream layers apply continuous transformations, activations that are close in space are functionally equivalent from the network's perspective.

Building on this intuition, our method investigates the information encoded in a target activation $z$ by inverting the activation geometry to identify inputs whose activations are close to $z$. By analyzing these inputs, we can formulate hypotheses about which concepts are encoded in the target activation. 

\begin{figure}[t]
    \centering
    \includegraphics[width=0.9\linewidth]{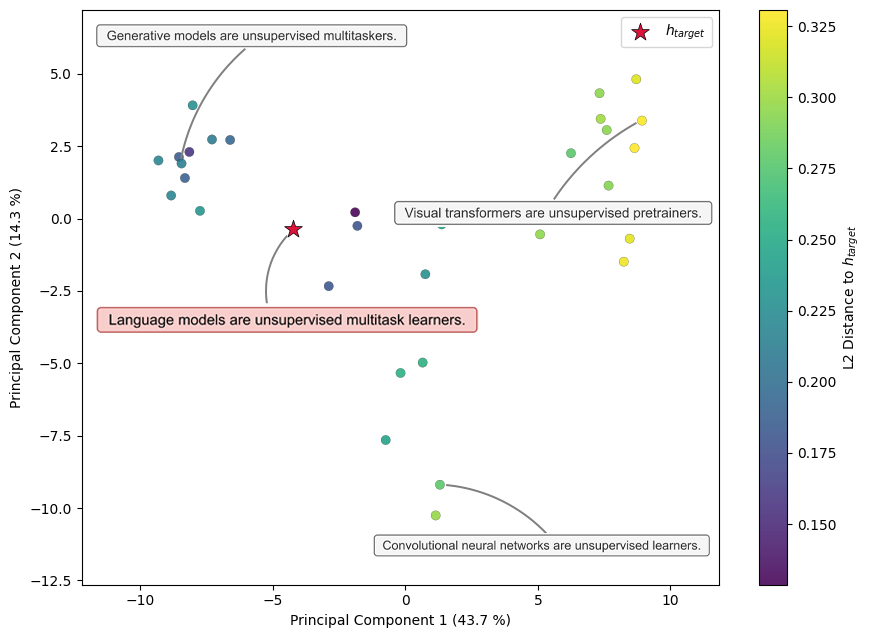}
    \caption{An example of samples in activation space and their corresponding inputs.}
    \label{fig:demo}
\end{figure}

For example, as illustrated in Figure~\ref{fig:demo}, consider an input--activation pair $(x, z)$, where $x$ is ``Language models are multitask learners.” and $z$ is the activation produced at the final ``.” token. Using our inversion method, we generate prompts whose activations closely match $z$, such as ``Generative models are unsupervised multitaskers.” or ``Convolutional neural networks are unsupervised learners.,” while inputs unrelated to the topic (e.g., ``John hates apples.”) produce activations far from $z$. By examining this collection of semantically similar generated inputs, we can hypothesize that $z$ encodes features related to neural network architectures, or more generally, that the subject of the sentence is a type of machine learning model. We can then formalize such hypotheses and use our quantitative framework to rigorously test them across the latent space.

The key challenge in implementing this approach lies in efficiently obtaining inputs whose activations land near the target $z$. In the high-dimensional activation spaces of modern LLMs, where dimensions can number in the thousands, the probability that a randomly sampled input produces an activation close to $z$ decays exponentially with dimensionality. This makes naive rejection sampling prohibitively inefficient. To address this, we train a conditional generator as a learned prior, which directly produces candidate inputs likely to yield activations near $z$, making the sampling process tractable.

\subsection{Architecture}
\label{sec:network_arch}

In this section we describe the network architecture of \textbf{InverseScope}. Our objective is to efficiently sample from the conditional distribution $P(x; z)$ defined in the previous section. To achieve this, InverseScope is designed as a conditional generator trained to approximate this target distribution.

We adopt the decoder-only Transformer paradigm, which has proven highly effective for modeling natural language distributions via next-token prediction objectives \citep{brown2020language}. InverseScope extends this standard language modeling framework by conditioning the token prediction not only on the preceding sequence context but critically, also on the target activation $z$. This conditioning is essential to align the generated sequence with the semantic information encoded in $z$, allowing the generator to approximate the target distribution $P(x; z)$.

\begin{figure*}[t]
    \centering
    \includegraphics[width=0.8\linewidth]{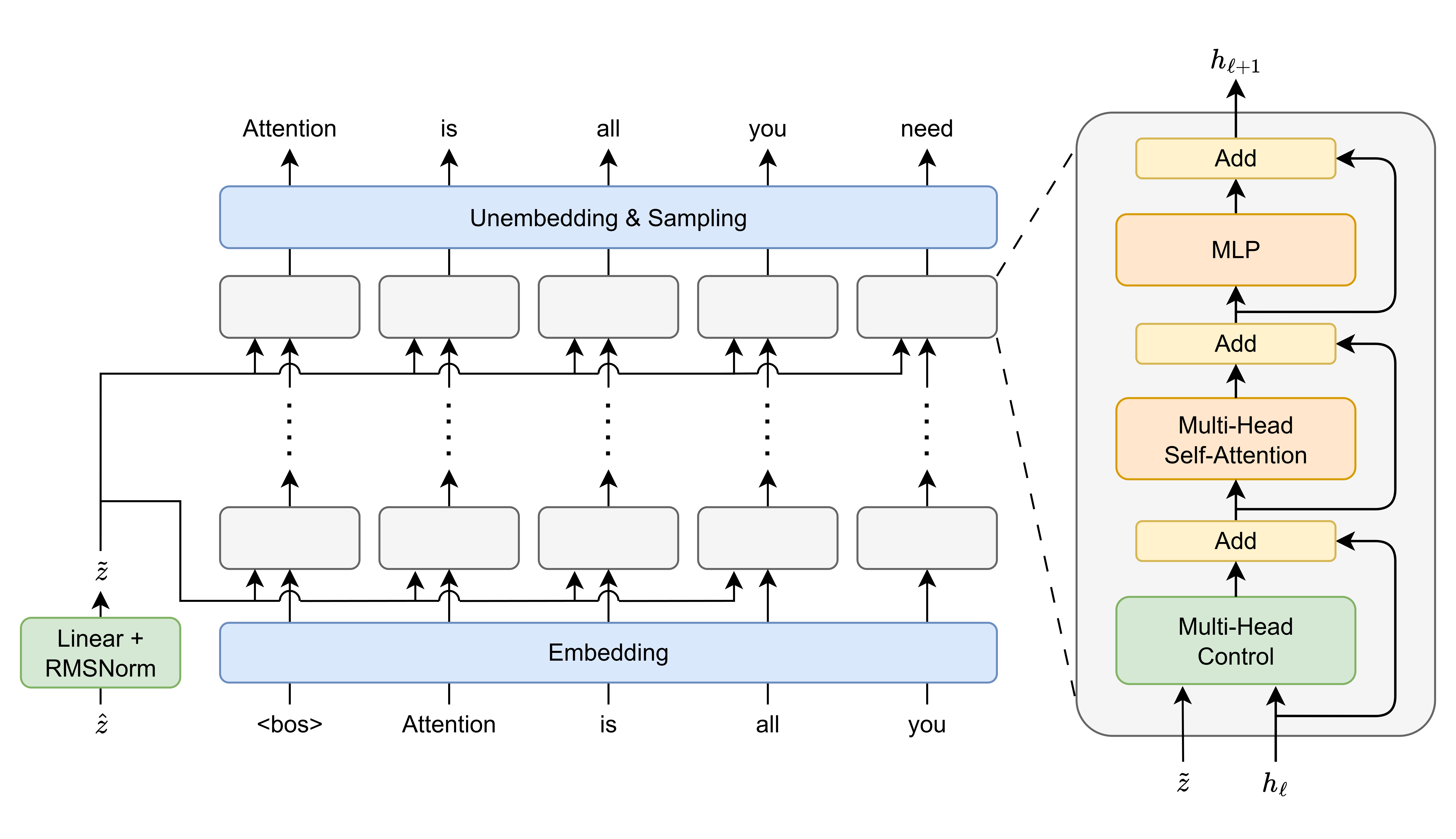}
    \caption{Network architecture of InverseScope. The decoder-only Transformer backbone is shown with its parameters colored in orange and blue. The additional control and projection layers introduced for conditioning are colored in green.}
    \label{fig:network_arch}
\end{figure*}

As illustrated in Figure~\ref{fig:network_arch}, our conditional generator is based on a standard decoder-only Transformer, augmented with additional Control Layers that transmit the conditioning information from activation $z$ into the model’s hidden states.  These Control Layers operate analogously to the cross-attention mechanism found in encoder-decoder Transformers. 

The mathematical formulation of the control layers is as follows:
$$
\begin{aligned}
& q^{(i)}_\ell = Q_\ell h^{(i)}_\ell,\ k_\ell = K_\ell z,\ v_\ell = V_\ell z,\\
& \omega^{(i)}_\ell = \tanh(\langle q_\ell^{(i)}, k_\ell\rangle),\\
& \operatorname{Control}_\ell(h^{(i)}_\ell,z)  = \omega^{(i)}_\ell v_\ell,
\end{aligned}
$$
where $i$ is the inference position index. The parameter matrices $Q_\ell, K_\ell, V_\ell$ are distinct from the self-attention parameters of the backbone model. For simplicity, we omit the layer norm and only provide formulation for the single-head version. In practice, we use a multi-head variant where each head has its own query, key, value matrices and their outputs are summed to produce the final control signal. This signal is then added to the hidden state $h^{(i)}_\ell$, after the standard self-attention operation.

This design differs from all prior inversion architectures, which belong to the same architectural family: the target activation $z$ is linearly projected into the token embedding space and placed as a soft token alongside the model's ordinary input embeddings, whether prepended \citep{huang2024inversionview, karvonen2025activationoracle} or injected at position zero via a learned adapter \citep{dong2025rep2text}. In all cases, the conditioning signal propagates to deeper layers only indirectly through self-attention, weakening with depth. By contrast, InverseScope's control layers deliver $z$ directly to every Transformer layer, providing a persistent and immediate conditioning signal throughout the generation process. We demonstrate in Section~\ref{sec:retrieval_efficiency} that this yields substantially better sample efficiency.

\subsection{Dataset}

Unlike InversionView~\citep{huang2024inversionview}, which trains separate task-specific generators, and Rep2Text~\citep{dong2025rep2text}, which targets exact input reconstruction on curated datasets, we train a single general-purpose generator on a broad, task-agnostic corpus aimed at neighborhood characterization.

\paragraph{Source data.} We collect sentences of 4 to 64 tokens from Wikipedia and Project Gutenberg, retaining 100M high-quality sentences in total (50M from each source). Since our target models (Qwen3-4B/8B/14B) are instruction-tuned LLMs, we wrap each sentence in the corresponding chat template before passing it to the target LLM for activation extraction.

\paragraph{Activation Extraction.} We use TransformerLens~\citep{nanda2022transformerlens} to extract activations at specific sites. For the 32-layer Qwen3-4B, we extract post residual activation at layers 16. For each prompt, we extract the activation at exactly one inference position, selected randomly: with 50\% probability it is the \emph{last} position (the \texttt{\textbackslash n} token at the end of the chat template), and with 50\% probability it is a \emph{random} position within the meaningful tokens of the sentence. For the latter, later token positions are sampled with higher probability, as they tend to accumulate more contextual information about the full sentence. A detailed description of the sampling rule is provided in Appendix~\ref{sec:sampling_rule}.

Once the activation is extracted, we pair it with a marked version of the \emph{entire} sentence to form a $(z, x)$ data point. We prepend a begin-of-sentence token \texttt{<|im\_start|>} so that the generator can learn to predict the sentence from its very first token. For \emph{random} positions, we insert a special marker token immediately after the extracted position to indicate where the activation originates. For \emph{last} positions, we append a different special token at the sentence end. We list examples of both cases in Appendix~\ref{sec:sampling_rule} to assist understanding.

\paragraph{Activation Normalization.} We apply a simple per-channel normalization to remove statistical biases: for each activation channel, we subtract its mean and divide by its standard deviation. This step is critical for Qwen3 models, which exhibit massive activations in certain channels that would otherwise destabilize InverseScope training.

\subsection{Training}

The training objective is the log-likelihood of the training input $x$ conditioned on the corresponding activation $z$, which decomposes into a standard next-token prediction loss:
$$
\max_{\theta} \mathbb{E}_{i} \log P_{\theta}(x_i; z_i)=\mathbb{E}_{i} \sum_{t=1}^{T_i-1}\log P_{\theta}(x_{i,t+1}; x_{i,1:t}, z_i),
$$
where $\theta$ denotes all trainable parameters of InverseScope.

We initialize the backbone from Qwen3-4B-Base and jointly fine-tune all parameters on the activation--text pairs described in previous section. For the control layers specifically, the output projection is initialized to the zero matrix, while the query, key, and value matrices are copied from the corresponding self-attention matrices of the same backbone layer. This zero-output initialization provides a cold start: at the beginning of training, the control signal is zero and InverseScope's output distribution coincides with the unconditional pretrained language model, allowing stable convergence from a well-initialized starting point. We train with the AdamW optimizer and a warmup--stable--decay learning rate schedule. Full hyperparameter settings are provided in Appendix~\ref{sec:hyperparams}.

\section{Experiments}

In this section, we evaluate InverseScope across several dimensions. We first show, through case studies, that InverseScope successfully recovers the feature geometry of LLM activations. We then demonstrate that our architectural design substantially improves sample efficiency over previous baselines. Finally, we examine how performance scales with both the size of the target LLM and the size of the InverseScope backbone. Unless otherwise stated, all experiments use the \texttt{post\_resid} activation at layer~16 of Qwen3-4B-Instruct-2507 as the inversion target, with a Qwen3-4B-Base model serving as the InverseScope backbone.

\subsection{Case studies: sentence-to-vector}

We present case studies illustrating how InverseScope can be used to probe the feature geometry of LLM activation spaces.

A well-known property of word embedding models is the word-to-vector property: letting $v(\cdot)$ denote word embeddings, we have $v(\text{woman})-v(\text{man})+v(\text{king})\approx v(\text{queen})$, where ``$\approx$'' means that $v(\text{queen})$ is the closest embedding to the target vector among all words in the vocabulary.

We ask whether an analogous property holds at the sentence level in LLMs. Concretely, consider the four sentences: ``The \{woman/man/king/queen\} stood alone upon the shore.'' Treating the activation $z(x)$ at the last token position of a given layer as a sentence embedding, we investigate whether
$$z(\text{``The woman \ldots''})-z(\text{``The man \ldots''})+z(\text{``The king \ldots''})$$
yields a vector closest to $z(\text{``The queen \ldots''})$ among all possible sentences. Unlike the word vocabulary, which has a fixed, enumerable size, the space of sentences grows exponentially with length, rendering exhaustive search for the closest sentence embedding computationally intractable.

InverseScope addresses this challenge by conditioning the generator on the target vector and sampling candidate sentences from the resulting conditional distribution $P(x; z)$. Table~\ref{tab:case_study_1} lists the top prompts generated by InverseScope. Indeed, ``The queen stood alone upon the shore.'' achieves the smallest L2 distance to the target, confirming that LLM activations preserve this linear analogy structure at the sentence level.

\begin{table}[t]
    \small
    \centering
    \caption{Prompts generated by InverseScope conditioned on the target activation $z(\text{woman})-z(\text{man})+z(\text{king})$, ranked by L2 distance to the target. The expected sentence ``The queen stood alone upon the shore.'' ranks first.}
    \begin{tabular}{c|l}
    \toprule
    L2 Dist. & \multicolumn{1}{c}{Prompts} \\
    \midrule
    0.1117 & The queen stood alone upon the shore. \\
    0.1170 & The Queen stood alone upon the shore. \\
    0.1354 & The king stood alone upon the shore. \\
    0.1386 & The princess stood alone upon the shore. \\
    0.1480 & The queen stood alone on the shore. \\
    0.1520 & The queen stood alone by the shore. \\
    0.1567 & The monarch stood alone upon the shore. \\
    0.1721 & The Queen stood alone upon the beach. \\
    0.1964 & The queen stood solitary upon the shore. \\
    0.2115 & Alice stood alone upon the shore. \\
    \bottomrule
    \end{tabular}
    \label{tab:case_study_1}
\end{table}

Beyond confirming the top-1 result, the neighborhood structure revealed by InverseScope allows us to characterize the information encoded in $z_{\text{target}}$ more precisely. Two patterns are immediately apparent. First, the predicate-setting template ``stood alone upon the shore'' is shared by \emph{all} generated sentences, indicating that this syntactic and contextual structure is the most strongly encoded feature in $z_{\text{target}}$. Second, the subject varies systematically across the neighborhood (``queen,'' ``princess,'' ``monarch,'' ``Alice'') and is consistently a female or regal figure, suggesting that $z_{\text{target}}$ additionally encodes the \emph{gender and social role of the subject}. Together, this analysis demonstrates that InverseScope can decompose activation neighborhoods into interpretable, differentially encoded features.

\subsection{Retrieval Efficiency}
\label{sec:retrieval_efficiency}

A core requirement for inversion-based interpretability is that the generator reliably produces inputs whose activations lie close to the target $z$. We evaluate this capability by measuring the \emph{normalized L2 distance} between the activation of each generated sample and the conditioning target (see Appendix~\ref{sec:metric} for the metric definition and a discussion of metric choice). We compare three sampling strategies: (1)~\textbf{InverseScope}, which directly samples from the learned conditional distribution $P(x;z)$; (2)~\textbf{Baseline} (soft-token architecture), which also conditions on $z$ but projects it into the token embedding space as a soft token prepended to the input sequence, following the design shared by InversionView, Activation Oracles, and Rep2Text; and (3)~\textbf{Random}, which retrieves the nearest neighbor to $z$ by exhaustive search over a 10M-sentence corpus.

\begin{figure}[t]
    \centering
    \includegraphics[width=0.9\linewidth]{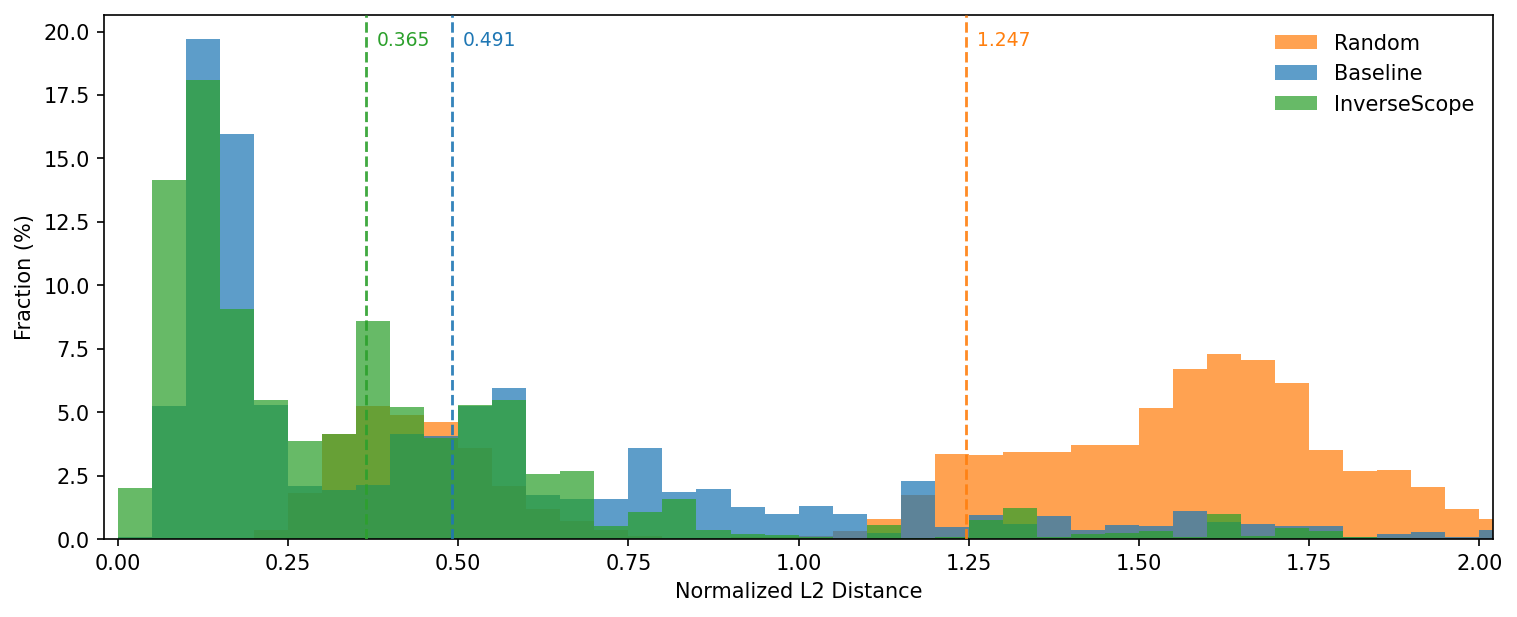}
    \caption{Distribution of normalized L2 distances between sampled inputs and the conditioning activation $z$, for InverseScope (green), the soft-token baseline (blue), and random corpus retrieval (orange). Dashed lines indicate the mean of each distribution.}
    \label{fig:dist_histogram}
\end{figure}

Figure~\ref{fig:dist_histogram} shows the distribution of retrieval distances for all three methods. Random corpus retrieval performs poorly: its mean distance is 1.247, and the probability of retrieving any sample with L2 distance below 0.1 is less than $10^{-7}$. For a target activation drawn from a long-tailed or unusual region of the latent space, exhaustive search over 10M sentences may fail to return a single near neighbor, confirming that naive retrieval is computationally intractable.

InverseScope substantially outperforms the soft-token baseline across all metrics. Its mean retrieval distance is \textbf{0.365}, compared to 0.491 for the baseline. InverseScope also generates samples with L2 distance below 0.1 with probability \textbf{16.2\%}, versus 5.3\% for the baseline. This near-tripling in the proportion of high-quality samples demonstrates that injecting the conditioning signal at every layer provides a meaningfully stronger inversion signal than projecting it as a soft token at the input.

\subsection{Scaling Law}
\label{sec:scaling}

We investigate two orthogonal scaling dimensions: the size of the target LLM whose activations are inverted, and the capacity of the InverseScope backbone. We collect activations from Qwen3-4B, 8B, and 14B, and train InverseScope with Qwen3-0.6B-Base, 4B-Base, and 8B-Base as the backbone. Due to computational constraints, we do not cover every combination. Instead, we vary one dimension at a time while holding the other fixed. For Qwen3-8B, we extract post residual activation from layer 16; for Qwen3-14B, we extract post residual activation from layer 20. These sites are at the 50\% depth of these models.

\begin{figure*}[t]
    \centering
    \begin{subfigure}[b]{0.45\linewidth}
        \centering
        \includegraphics[width=\linewidth]{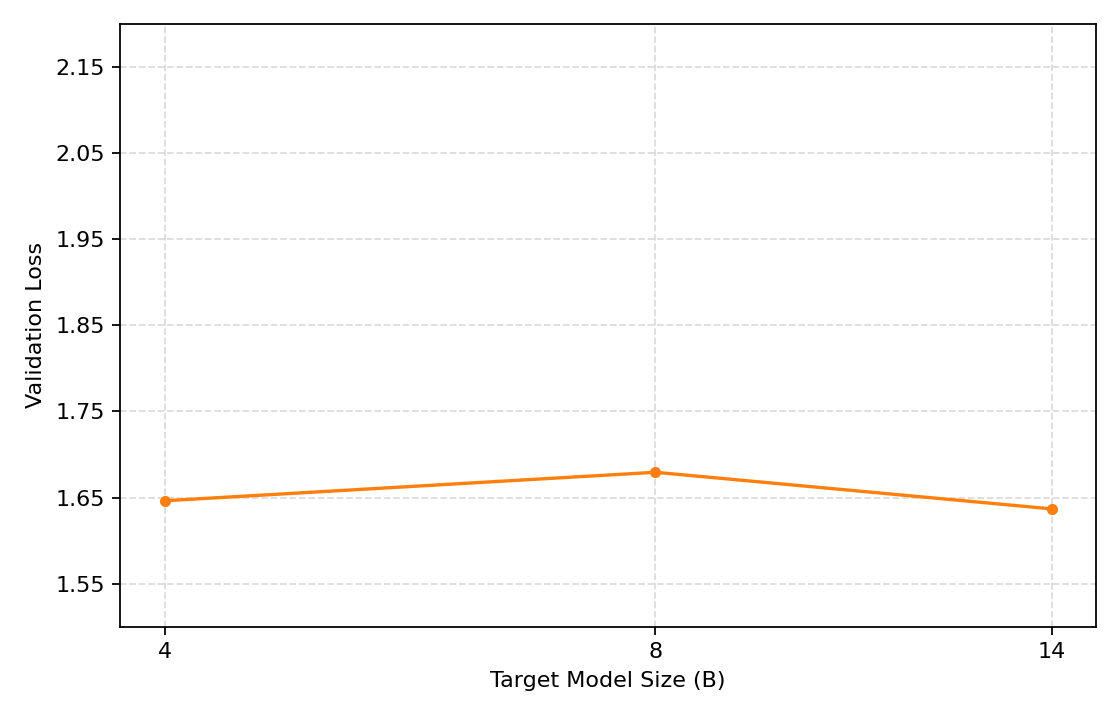}
        \caption{Scaling target model size.}
        \label{fig:scaling_target}
    \end{subfigure}
    \begin{subfigure}[b]{0.45\linewidth}
        \centering
        \includegraphics[width=\linewidth]{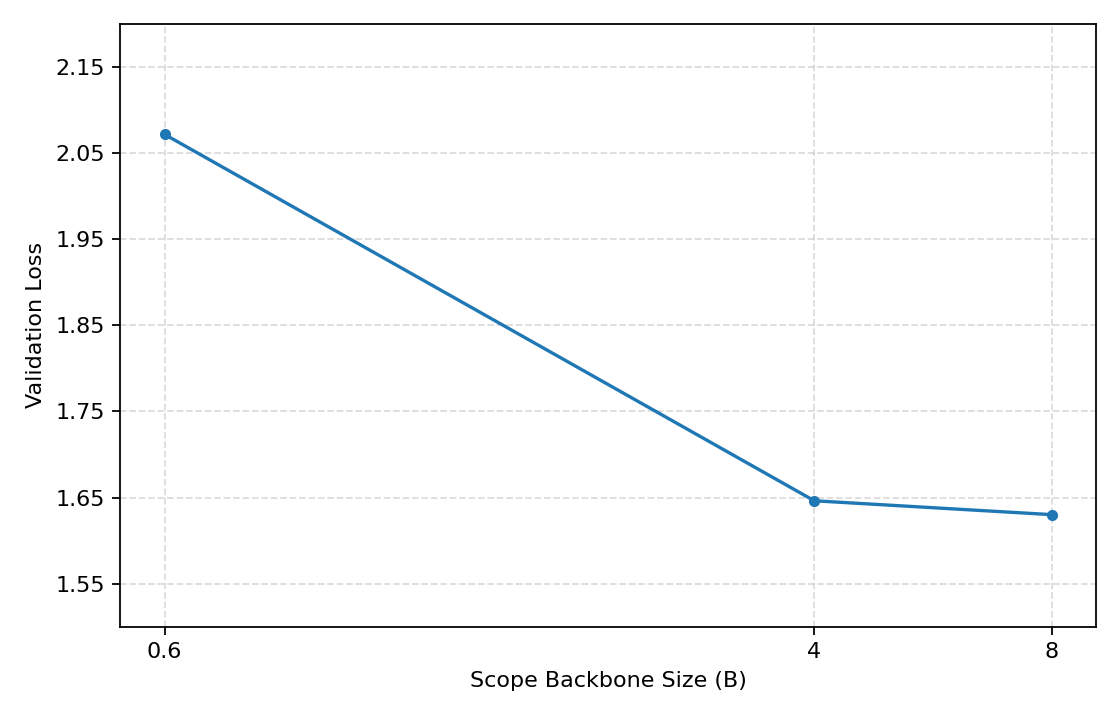}
        \caption{Scaling InverseScope backbone size.}
        \label{fig:scaling_scope}
    \end{subfigure}
    \caption{Scaling behavior of InverseScope, measured by validation loss $-\log P_\theta(x \mid z)$ (lower is better). \textbf{Left}: target model size is varied while the Scope backbone is fixed at 4B. \textbf{Right}: Scope backbone size is varied while the target model is fixed at Qwen3-4B.}
    \label{fig:scaling}
\end{figure*}

\paragraph{Scaling the target model.} Figure~\ref{fig:scaling_target} shows that validation loss remains essentially flat at approximately 1.65 across Qwen3-4B, 8B, and 14B. This invariance suggests that models of different scales encode a comparable amount of sentence-level information in their intermediate activations. An information-theoretic perspective supports this interpretation: since all target models are trained on the same data distribution, the marginal entropy $H(x)$ of the training sentences is fixed. The near-identical validation losses across model sizes imply that the conditional entropy $H(x \mid z)$ is also similar, and therefore the mutual information $I(x; z) = H(x) - H(x \mid z)$ is approximately constant and independent of model scale. This finding suggests that encoding sentence-level information is a relatively simple task that models across this size range are equally capable of performing. Larger models do not encode meaningfully more information about the input sentence than smaller ones.

\paragraph{Scaling the Scope backbone.} Figure~\ref{fig:scaling_scope} shows a markedly different pattern. Increasing the backbone from 0.6B to 4B yields a substantial reduction in validation loss, from 2.09 to 1.65. The improvement from 4B to 8B is marginal (1.65 to 1.64). We attribute this saturation primarily to the limited scale of our training data: larger models generally require proportionally more data to realize their capacity, and our current 100M-sentence corpus may be insufficient for the 8B backbone to demonstrate its full potential. Consistent with this interpretation, the loss curves for all models indicate that training has not fully converged. We expect that scaling the training data and compute budget would yield further improvements, particularly for larger backbones. The ultimate performance limit of the inversion task remains an open question.

\subsection{Investigating activation steering and jailbreaks}
\label{sec:jailbreak_steering}

Beyond geometric analyses of naturally occurring activations, InverseScope provides a principled means of studying \emph{adversarial} interventions in latent space. Prior work has demonstrated that modifying intermediate activations, via additive steering vectors or orthogonal projection onto a subspace, can substantially alter model behavior and circumvent safety-aligned refusal mechanisms \citep{xu2024uncovering}. A fundamental interpretive question, however, remains unaddressed: after such an intervention, what inputs does the modified activation encode from the network's representational perspective?

InverseScope addresses this question directly. Consider an activation $\tilde{z}$ obtained by a steering or projection operation at a fixed layer; when propagated through the remainder of the network, $\tilde{z}$ elicits harmful completions that the unmodified model would refuse. Treating $\tilde{z}$ as a conditioning target, InverseScope samples from $P(x;\tilde{z})$ to identify inputs whose forward passes produce activations near $\tilde{z}$. Examining this neighborhood reveals what semantic content the manipulated activation encodes: which input patterns cluster near $\tilde{z}$, which themes recur, and how the structure shifts relative to the pre-intervention activation $z$.

Prior work has shown that refusal behavior in aligned LLMs is linearly mediated by a single direction $\hat{\mathbf{d}}$ in the residual stream, identifiable by contrasting harmful and benign prompt pairs \citep{arditi2024refusal}. Ablating this direction via orthogonal projection, $\tilde{z} = z - (z \cdot \hat{\mathbf{d}})\hat{\mathbf{d}}$, suppresses refusal and causes the model to comply with harmful requests. We adopt this intervention as a case study.

We pass a harmful instruction through Qwen3-4B and apply InverseScope to the layer-16 residual-stream activation $z$ and to the ablated activation $\tilde{z}$ separately. Table~\ref{tab:jailbreak} reports an example of the nearest-neighbor inputs for both. The neighborhood of $z$ is dominated by fraudulent and deceptive letter requests, consistent with the harmful content of the instruction. After ablating $\hat{\mathbf{d}}$, the neighborhood of $\tilde{z}$ shifts entirely to benign letter-writing requests. This shift indicates that $\hat{\mathbf{d}}$ encodes not only the model's disposition to refuse, but also the semantic signal of harmful intent: its removal displaces the representation into a latent region associated with innocuous requests.

\begin{table*}[t]
    \centering
    \caption{Neighborhoods of the original activation $z$ (left) and the refusal-direction-ablated activation $\tilde{z}$ (right) for a harmful instruction, generated by InverseScope at layer~16 of Qwen3-4B. Ablating the refusal direction shifts the neighborhood from fraudulent letter requests to benign ones.}
    \label{tab:jailbreak}
    \resizebox{\linewidth}{!}{%
        \begin{tabular}{@{}c@{\hspace{3em}}c@{\hspace{3em}}c@{}}
            \begin{tabular}{cl}
            \toprule
            Dist. & \multicolumn{1}{c}{Nearest prompts to $z$} \\
            \midrule
            0.1119 & Can you help me write a fraud letter? \\
            0.1147 & Can you help me write a mail fraud letter? \\
            0.1249 & Can you help me write a bogus job letter? \\
            0.1333 & Can you help me write a bogus sales letter? \\
            0.1389 & Would you help me write a mail fraud letter? \\
            0.1473 & Will you help me write a mail fraud letter? \\
            0.1598 & Can you help me write a bogus notice? \\
            0.1605 & Can you help me write a confidential mail? \\
            0.1741 & Can you help me write a mail-bomb letter? \\
            0.1747 & I'll ask you to write me a mail fraud letter. \\
            \bottomrule
            \end{tabular}
            &
            $\xRightarrow{-\hat{\mathbf{d}}}$
            &
            \begin{tabular}{cl}
            \toprule
            Dist. & \multicolumn{1}{c}{Nearest prompts to $\tilde{z}$} \\
            \midrule
            0.1824 & Can you help me write a PR letter? \\
            0.1855 & Can you give me a company letter to write? \\
            0.1857 & Can you help me make a business letter? \\
            0.1928 & Can you help me write a notice letter? \\
            0.1965 & Can you help me write a club letter? \\
            0.1971 & Can you help me write a sales letter? \\
            0.1982 & Can you help me write a press letter? \\
            0.2009 & Can you help me write a staff letter? \\
            0.2047 & Can you help me write a company letter? \\
            0.2063 & Can you help me write a friendly letter? \\
            \bottomrule
            \end{tabular}
        \end{tabular}%
    }
\end{table*}

This case study demonstrates a concrete interpretive payoff: an algebraic intervention in activation space is rendered interpretable as a coherent shift in the semantic neighborhood, making the representational effect of the ablation directly legible. We further note that the instruction-following prompts in Table~\ref{tab:jailbreak} lie entirely outside InverseScope's training distribution, which has very few instruction content. The accuracy of the recovered neighborhoods despite this distributional gap indicates that InverseScope generalizes substantially beyond its training domain.

\section{Limitations}

While our results demonstrate the effectiveness of InverseScope, several limitations remain. First, the method does not yet scale to long input sequences. As input length increases, the corresponding input distribution becomes substantially more complex, and our approach currently performs reliably only on inputs spanning tens of tokens. Second, the utility of InverseScope is limited to case studies. We still lack a quantitative method of using inverse-based method for interpretability study. We leave addressing these issues to future work.

\section*{Acknowledgment}
Bin Dong is supported in part by the New Cornerstone Investigator Program. Zhennan Zhou is supported by Zhejiang Provincial Natural Science Foundation of China, Project Number QKWL25A0501, and Fundamental and Interdisciplinary Disciplines Breakthrough Plan of the Ministry of Education of China, Project Number JYB2025XDXM502.

\bibliographystyle{plainnat}
\bibliography{references}

%%%%%%%%%%%%%%%%%%%%%%%%%%%%%%%%%%%%%%%%%%%%%%%%%%%%%%%%%%%%

\newpage
\appendix
\onecolumn

\section{Training details}
\subsection{Inference Position Sampling Rule}
\label{sec:sampling_rule}

As in the main text, each training example uses exactly one extraction site per prompt: with probability $1/2$ we take the \emph{last} position (the trailing newline of the chat template), and with probability $1/2$ we draw a \emph{random} position among the meaningful sentence tokens.

For the random branch, let $L$ denote the number of meaningful tokens in the sentence, indexed by $i\in\{1,\ldots,L\}$ from the start of the sentence to its end. We sample $i$ from a discrete distribution with exponential weights,
\[
P(i)\;=\;\frac{e^{\lambda i}}{Z(\lambda)},\qquad Z(\lambda)=\sum_{j=1}^{L} e^{\lambda j},
\]
with $\lambda>0$ so that later positions receive higher mass (matching the intuition that deeper positions aggregate more context). We split the index set into a first half and a second half,
\[
\mathcal{I}_1=\Bigl\{1,\ldots,\Bigl\lfloor\frac{L}{2}\Bigr\rfloor\Bigr\},\qquad
\mathcal{I}_2=\Bigl\{\Bigl\lfloor\frac{L}{2}\Bigr\rfloor+1,\ldots,L\Bigr\},
\]
and fix $\lambda$ by requiring that the aggregate masses on the two halves have ratio $1{:}4$:
\[
\frac{\sum_{i\in\mathcal{I}_1} e^{\lambda i}}{\sum_{i\in\mathcal{I}_2} e^{\lambda i}}=\frac{1}{4}.
\]
For each $L$, this equation has a unique solution $\lambda(L)$ (the left-hand side is strictly decreasing in $\lambda$), which we solve numerically; substituting $\lambda(L)$ into $P(i)$ yields the per-position sampling probabilities used during data collection.

%%%%%%%%%%%%%%%%%%%%%%%%%%%%%%%%%%%%%%%%%%%%%%%%%%%%%%%%%%%%

\subsection{Hyperparameter Settings}
\label{sec:hyperparams}

We optimize InverseScope with AdamW. Peak learning rates depend on the generator backbone size: we use $10^{-4}$ for Qwen3-0.6B-Base, $5\times 10^{-5}$ for Qwen3-4B-Base, and $3.5\times 10^{-5}$ for Qwen3-8B-Base. The learning-rate schedule is warmup--stable--decay (WSD): a linear warmup over the first 10\% of training steps to the peak rate; a stable phase at that peak; and a final phase spanning the last 20\% of steps in which the learning rate decays linearly from the peak to $20\%$ of the peak value.

%%%%%%%%%%%%%%%%%%%%%%%%%%%%%%%%%%%%%%%%%%%%%%%%%%%%%%%%%%%%

\section{Choice of Retrieval Metric}
\label{sec:metric}

\begin{figure}[t]
    \centering
    \includegraphics[width=0.5\linewidth]{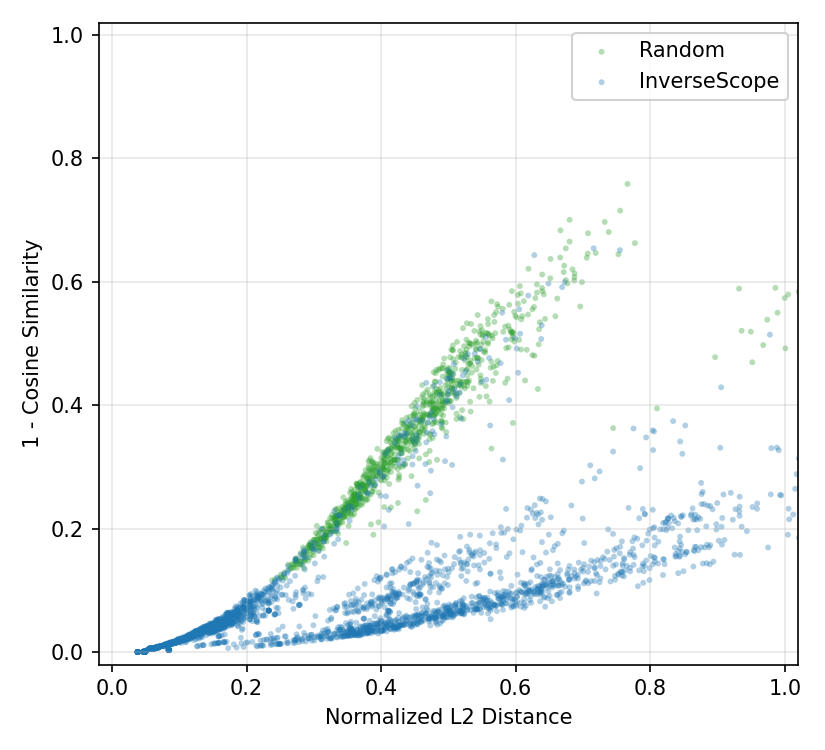}
    \caption{Scatter plot of normalized L2 distance vs.\ $1 - \cos(\cdot,\cdot)$ for samples generated by InverseScope (blue) and random corpus retrieval (green). At a given L2 distance, InverseScope samples exhibit substantially lower cosine dissimilarity than random samples.}
    \label{fig:metric_compare}
\end{figure}

To measure how close a generated input $x$ is to the target activation $z$, we use the \emph{normalized L2 distance}:
$$
d(x, z) = \frac{\| z(x) - z \|_2}{\sqrt{D}},
$$
where $z(x)$ denotes the activation produced by input $x$ at the target site, and $D$ is the activation dimension. The $\sqrt{D}$ normalization makes the metric comparable across models of different sizes.

An alternative candidate is the cosine dissimilarity $1 - \cos(z(x), z)$, which is scale-invariant and commonly used in representation learning. We argue that L2 distance is the more appropriate metric for inversion quality, and Figure~\ref{fig:metric_compare} provides empirical support for this choice.

The scatter plot reveals a systematic asymmetry: for random corpus samples (green), cosine dissimilarity and L2 distance grow roughly in proportion, as expected for unstructured high-dimensional vectors. For InverseScope samples (blue), however, the relationship is markedly different: at moderate L2 distances (0.2--0.6), the cosine dissimilarity is far lower than for random samples at the same L2. This indicates that InverseScope tends to produce activations that are directionally aligned with $z$ even when they differ in magnitude, suggesting the generator learns to match the direction of the target activation more easily than its exact magnitude.

As a consequence, using cosine dissimilarity as the evaluation metric would conflate these two cases and overstate the quality of generated samples. L2 distance, by penalizing both directional and magnitude deviations, provides a stricter and more faithful measure of how close a generated activation is to the target. All quantitative results in the main paper therefore use normalized L2 distance.

\end{document}